\documentclass[conference]{IEEEtran}
\IEEEoverridecommandlockouts
\usepackage{amsmath,amssymb,amsfonts}
\usepackage{algorithmic}
\usepackage{graphicx}
\usepackage{tabularx}
\usepackage{textcomp}
\usepackage{xcolor}
\usepackage{booktabs}
\usepackage{enumitem}
\usepackage[numbers]{natbib}

\def\BibTeX{{\rm B\kern-.05em{\sc i\kern-.025em b}\kern-.08em
    T\kern-.1667em\lower.7ex\hbox{E}\kern-.125emX}}

\PassOptionsToPackage{hyphens}{url}    
\usepackage{tikz,xcolor,hyperref}
\usepackage{microtype}
\usetikzlibrary{shapes,arrows,positioning}

\definecolor{lime}{HTML}{A6CE39}
\definecolor{prpa}{HTML}{648fff}
\definecolor{prpb}{HTML}{785ef0}
\definecolor{prpc}{HTML}{dc267f}
\definecolor{prpd}{HTML}{fe6100}
\definecolor{prpe}{HTML}{ffb000}

\DeclareRobustCommand{\orcidicon}{
	\begin{tikzpicture}
	\draw[lime, fill=lime] (0,0) 
	circle [radius=0.16] 
	node[white] {{\fontfamily{qag}\selectfont \tiny ID}};
	\draw[white, fill=white] (-0.0625,0.095) 
	circle [radius=0.007];
	\end{tikzpicture}
	\hspace{-2mm}
}
\foreach \x in {A, ..., Z}{%
	\expandafter\xdef\csname orcid\x\endcsname{\noexpand\href{https://orcid.org/\csname orcidauthor\x\endcsname}{\noexpand\orcidicon}}
}

\begin{document}

\title{Detecting Hallucinations in Large Language Model Generation: A Token Probability Approach}
\author{
\IEEEauthorblockN{Ernesto Quevedo\orcidD{}}
\IEEEauthorblockA{Department of Computer Science\\
School of Eng. \& Computer Science\\
Baylor University\\
Email: Ernesto\_Quevedo1@Baylor.edu}
\and
\IEEEauthorblockN{Jorge Yero Salazar\orcidE{}}
\IEEEauthorblockA{Department of Computer Science\\
School of Eng. \& Computer Science\\
Baylor University\\
Email: Jorge\_Yero1@Baylor.edu}
\and
\IEEEauthorblockN{Rachel Koerner\orcidF{}}
\IEEEauthorblockA{Department of Computer Science\\
School of Eng. \& Computer Science\\
Baylor University\\
Email: Rachel\_Koerner1@Baylor.edu}
\and
\IEEEauthorblockN{Pablo Rivas\orcidC{}, \emph{Senior, IEEE}}
\IEEEauthorblockA{Department of Computer Science\\
School of Engineering \& Computer Science\\
Baylor University\\
Email: Pablo\_Rivas@Baylor.edu}
\and
\IEEEauthorblockN{Tomas Cerny\orcidG{}}
\IEEEauthorblockA{Department of Systems \& Industrial Engineering\\
College of Engineering\\
The University of Arizona\\
Email: Tomas\_Cerny@Baylor.edu}
}

\maketitle

\begin{abstract}
Concerns regarding the propensity of Large Language Models (LLMs) to produce inaccurate outputs, also known as hallucinations, have escalated. Detecting them is vital for ensuring the reliability of applications relying on LLM-generated content. Current methods often demand substantial resources and rely on extensive LLMs or employ supervised learning with multidimensional features or intricate linguistic and semantic analyses difficult to reproduce and largely depend on using the same LLM that hallucinated. This paper introduces a supervised learning approach employing two simple classifiers utilizing only four numerical features derived from tokens and vocabulary probabilities obtained from other LLM evaluators, which are not necessarily the same. The method yields promising results, surpassing state-of-the-art outcomes in multiple tasks across three different benchmarks. Additionally, we provide a comprehensive examination of the strengths and weaknesses of our approach, highlighting the significance of the features utilized and the LLM employed as an evaluator. We have released our code publicly at \url{https://github.com/Baylor-AI/HalluDetect}.
\end{abstract}

\begin{IEEEkeywords}
Large Language Models, Hallucinations
\end{IEEEkeywords}

\section{Introduction}
Large Language Models (LLMs) have become the core of many state-of-the-art Natural Language Processing (NLP) algorithms and have revolutionized various domains in NLP and computer vision and even more specialized applications in healthcare, finance, and the creative arts. Because of their impressive Natural Language Generation (NLG) capabilities~\cite{zhao2023survey, kaddour2023challenges}, they have attracted great interest from the public with great modern tools like ChatGPT~\cite{hosseini2023exploratory}, Github-Copilot~\cite{chen2021evaluating}, Dalle~\cite{zeqiang2023mini}, and others~\cite{zhao2023survey}. These models, with millions to billions of parameters, are often praised for their impressive ability to generate human-like text and tackle intricate tasks with limited to no fine-tuning with techniques like In-Context-Learning~\cite{lu2023emergent}.

Since many of the most popular applications and state-of-the-art algorithms in NLP rely on LLMs, any error they produce affects the results. Particularly in the cases of a Chatbot like ChatGPT, the generated responses are expected to maintain factual consistency with the source text~\cite{lei2023chain}. Currently, a pressing concern with LLMs is their propensity to ``hallucinate," which intuitively means to produce outputs that, while seemingly coherent, might be misleading, fictitious, or not genuinely reflective of their training data or actual facts~\cite{ji2023survey}. 


Furthermore, the consequences of hallucinatory-generated text when used by the public are a significant ethical concern. This fictitious content can lead to misinformation and have severe implications in delicate medical, legal, educational, and financial fields. Besides the ethical consequences, these errors can lead to limitations in the use of the LLMs to automate programming tasks completely and tedious hand-work, limiting their contribution to NLP tasks~\cite{ji2023survey,kaddour2023challenges}.


While there have been efforts to detect and mitigate hallucinations, many of the prevalent methods rely on supervised learning with many multidimensional features. In contrast, others used in-context-learning techniques based on intricate linguistic and semantic analyses~\cite{zhang2023siren,manakuletal2023selfcheckgpt,lei2023chain,DBLP:conf/iclr/0002WSLCNCZ23}. These methods affect the latency for use in real time. Additionally, current research has shown that even state-of-the-art approaches~\cite{ji2023survey, kaddour2023challenges, li-etal-2023-halueval, lei2023chain} struggle to detect hallucinations. 



However, recent research has hinted at the potential of numerical features mathematically~\cite{lee2023mathematical} and empirically~\cite{manakuletal2023selfcheckgpt,azariamitchell2023internal,su2024unsupervised}, as indicators of hallucinations on LLM outputs. These features could provide a resource-efficient method to detect and mitigate hallucinations. In this paper, we introduce a supervised learning approach employing two classifiers that use four numerical features derived from tokens and vocabulary probabilities obtained from other LLM evaluators, usually different ones.


Our research not only highlights the effectiveness of this method in comparison with current approaches in some scenarios but also paves the way for potential uses that validate the credibility of LLM outputs. Our main contributions are:








\begin{itemize}[noitemsep]
    \item Propose a supervised learning approach using four features to detect hallucinations in conditional text generated by LLMs, achieving success with two classifiers.

    \item Evaluate the performance of this approach across three datasets, comparing it with state-of-the-art methods and highlighting its strengths and weaknesses.

    \item Explore the impact of using the same LLM-Generator vs. different LLMs as evaluators, finding that alternative LLMs provide better indicators to identify hallucinations.

    \item Compare the difference in performance when using smaller LLM evaluators concerning bigger LLM evaluators like LLaMa-Chat-7b~\cite{touvron2023llama}.

    \item Feature importance study with ablation and coefficients analysis.
\end{itemize}


This paper is structured as follows. First, we present the related works. Second, we describe our methodology. Next, we offer the experiments performed and the results in three datasets. After that, we present an ablation analysis, followed by the conclusions. Finally, we conclude the paper with the limitations section.

\section{Related Work}
The occurrence of hallucinations in LLMs raises concerns, compromising performance in practical implementations like chatbots producing incorrect information. Various research directions have been explored to detect and mitigate hallucinations in different Natural Language Generation tasks~\cite{ji2023survey}. A text summarization verification system has been proposed to detect and mitigate inaccuracies~\cite{zhaoetal2020reducing,ji2023survey}. In dialogue generation, hallucinations have been studied with retrieval augmentation methods~\cite{shusteretal2021retrievalaugmentation,ji2023survey}. Also,  researchers aim to understand why hallucinations occur in different tasks and how these reasons might be connected~\cite{zheng2023does,dasetal2022diving}.

Recent approaches to detect and mitigate hallucinations include self-evaluation~\cite{yinetal2023large} and self-consistency decoding for intricate reasoning tasks~\cite{DBLP:conf/iclr/0002WSLCNCZ23}. Knowledge graphs are proposed for gathering evidence~\cite{jiangetal2023structgpt}. Token probabilities as an indicator of model certainty have been used, addressing uncertainty in sequential generation tasks~\cite{xiaowang2021hallucination, malinin2020uncertainty}. Scores from conditional language models are used to assess text characteristics~\cite{yuan2021bartscore, fu2023gptscore}. 

Additionally, Azaria et al.~\cite{azariamitchell2023internal}  trained a classifier that outputs the probability that a statement is truthful based on the hidden layer activations of
the LLM as it reads or generates the statement. Recently, the work SelfCheckGPT suggests that LLM's probabilities correlate with factuality~\cite{manakuletal2023selfcheckgpt}. Furthermore, Su et al.~\cite{su2024unsupervised} introduced Modeling of Internal
states for hallucination Detection (MIND), an unsupervised training framework that leverages the internal states of LLMs for real-time hallucination detection without requiring manual
annotations. Finally, a mathematical investigation by Lee et al.~\cite{lee2023mathematical} suggests that token probabilities are crucial in generating hallucinations in GPT models under certain assumptions.




\begin{figure*}[h]
    \small
    \includegraphics[width=\textwidth]{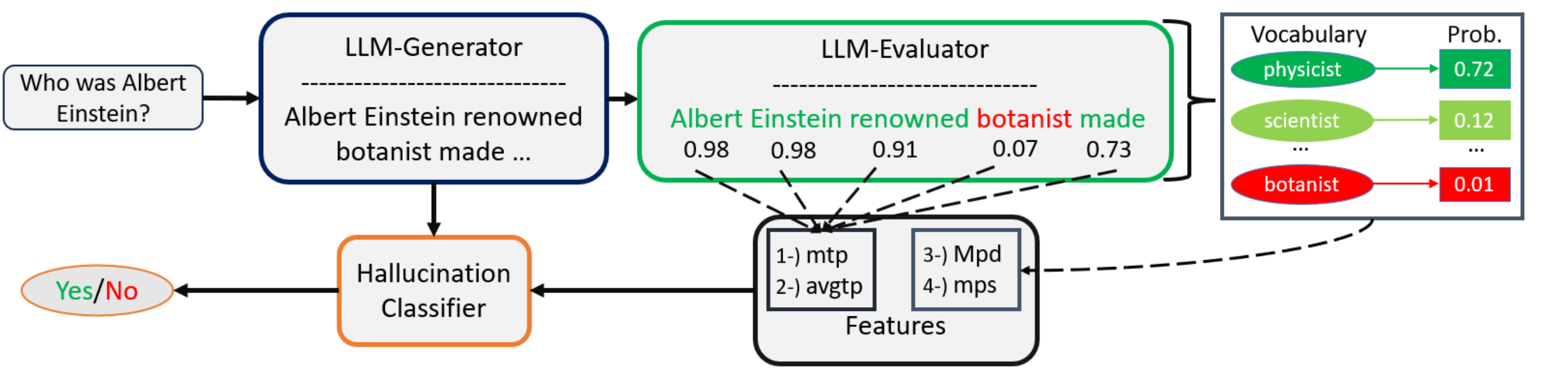}
    \caption{General Pipeline of the Proposed Methodology.}\label{fig:pipeline}
\end{figure*}


\section{Methodology} 
We implement two classifiers, a Logistic Regression (LR) and a Simple Neural Network (SNN), using four numerical features obtained from the token and vocabulary probabilities from a forward pass to an LLM with the conditional generation approach~\cite{10.1145/3617680}. In this section, we described our entire methodology to detect hallucinations on a generated text by an LLM conditioned on a piece of text.

\subsection{Problem Statement}
Given a pair of texts \emph{(condition-text, generated-text)} that represent the text used to condition the LLM to its generation. We want to detect if a given~\emph{generated-text} is a hallucination.

\subsection{General Pipeline}
Given a set of pairs of texts of the type \emph{(condition-text, generated-text)} from an LLM (we will call it the LLM-Generator ($LLM_G$)), we extract four numerical features based on the generated tokens and vocabulary tokens probabilities from another LLM that we call the LLM-Evaluator ($LLM_E$).\footnote{Which could be the same as $LLM_G$.} 


Then, using these four features, we trained two different classifiers: a Logistic Regression (LR) and a Simple Neural Network (SNN). Finally, we evaluate these classifiers on a test set they did not see before. Figure~\ref{fig:pipeline} illustrates the process.

\subsection{Features Description}
This section provides more details on each feature extracted. Every feature is computed using token probabilities and the vocabulary probability distribution corresponding to each token on the \emph{generated-text}. These can be formally defined as follows: (i) The token probability of each token of the Vocabulary of the $LLM_E$ corresponding to $t_i$ as $P_{LLM_{E_i}}(v_k) = (v_k | t_{i-1}, ..., t_1, c_{m}, ..., c_1; \theta)$ for every $k$. (ii) The token with the highest probability at position $i$ in the \emph{generated-text} according to $LLM_E$ as the $v^{*} = \arg \max_{k} P_{LLM_{E_i}}(v_k)$. (iii) The token with the lowest probability at position $i$ according to $LLM_E$ as the $v^{-} = \arg \min_{k} P_{LLM_{E_i}}(v_k)$.


Next is a natural language description of the four features and, the mathematical definition:

\begin{itemize}[noitemsep]
    \item \textbf{Minimum Token Probability (mtp)}: Minimum of the probabilities that the $LLM_E$ gives to the tokens on the \emph{generated-text}.
    \item \textbf{Average Token Probability (avgtp)}: Average of the probabilities that the $LLM_E$ gives to the tokens on the \emph{generated-text}.
    \item \textbf{Maximum $LLM_E$ Probability Deviation (Mpd)}: Maximum from all the differences between the token with the highest probability according to $LLM_E$ at position $i$ and the assigned probability from $LLM_E$ to $t_i$ which is the token generated by $LLM_G$.
    \item \textbf{Minimum $LLM_E$ Probability Spread (mps)}: Maximum from all the differences between the token with the highest probability according to $LLM_E$ at position $i$ ($v^*$) and the token with the lowest probability according to $LLM_E$ at position $i$ ($v^-$).
\end{itemize}

Formally, these features can be defined as:
\begin{align*}
    mtp &= \min_{i} P_{LLM_E}(t_i) ~~~ avgtp = \frac{\sum_{i = 1}^{n}{P_{LLM_E}(t_i)}}{n}\\
    Mpd &= \max_{1\le i \le n} (P_{LLM_E}(v^{*}) - P_{LLM_E}(t_i))  \\
    mps &= \min_{1\le i \le n} (P_{LLM_{E_i}}(v^{*}) - P_{LLM_{E_i}}(v^{-})) 
\end{align*}





These four numerical features are inspired by the mathematical investigation of the GPT model in \cite{lee2023mathematical}, and recent results in \cite{manakuletal2023selfcheckgpt,su2024unsupervised}, suggesting there is a correlation between the minimum token probability on the generation, the average of the token probabilities, and the average and maximum entropy. 


Lee et al.~\cite{lee2023mathematical} proposes that a reliable indicator of hallucination during GPT model generation is the low probability of a token being generated. This is based on the assumption that forcing the model to generate such a low-probability token occurs when the difference between the token with the highest probability and all other tokens is less than a small constant $\delta$. Here, $mps$ is an estimator to avoid the cost of calculating differences across a large vocabulary and generated text.

Additionally, Azaria et al.~\cite{azariamitchell2023internal}  trained a simple classifier called  Statement Accuracy Prediction, based on Language Model Activations (SAPLMA) that outputs the probability that a statement is truthful, based on the hidden layer activations of the LLM as it reads or generates the statement. The authors showed that SAPLMA, which leverages the LLM’s internal states, performs better than prompting the LLM to state explicitly whether a statement is true or false. Different from~\cite{azariamitchell2023internal},  Su et al.~\cite{su2024unsupervised} introduced MIND, an unsupervised training framework that leverages the internal states of LLMs for hallucination detection requiring no manual annotations. 

\subsection{Distinctive Methodological Approach}

Diverging from these three previous papers~\cite{azariamitchell2023internal,lee2023mathematical,manakuletal2023selfcheckgpt,su2024unsupervised}, the approach here differs in several aspects. This is an empirical paper, not a theoretical one like~\cite {lee2023mathematical}. We do not use Self-Consistency as~\cite{manakuletal2023selfcheckgpt} does, and also, our approach is not zero-shot or few-shot learning. Our approach follows supervised learning like~\cite{azariamitchell2023internal}. However, instead of using the contextual embeddings and hidden layers, we only use four features that result from aggregations of the token and vocabulary probabilities. We also test a simpler Logistic Regression classifier besides a Simple Dense Neural network.

Moreover, instead of using only the LLM generating the text ($LLM_G$), the argument is made that depending on the task and model type, different LLM-Evaluators ($LLM_E$) can provide consistent but quantitatively different results than using probabilities from $LLM_G$. The belief is that probabilities from a different model, varying in architecture, size, parameters, context length, and training data, can also serve as reliable indicators of hallucinations in the text generated by $LLM_G$. Since $LLM_E$ and $LLM_G$ are not always the same, an additional numerical feature, $Mpd$ (Maximum $LLM_E$ Probability Deviation), is introduced. This feature indicates the difference between the maximum probability token in the vocabulary of $LLM_E$ and the token generated by $LLM_G$.




Using different LLMs as evaluators takes advantage of the diversity of training data among different language models (LLMs) and captures various linguistic patterns and styles. Detecting hallucinations in LLM outputs, such as in $LLM_G$, might be possible through analyzing probability distributions. Yet, specific patterns may remain undetectable, potentially addressed by other models specialized in particular topics. Evaluations from multiple models enhance robustness by mitigating biases inherent in individual models' training data.

\subsection{Feature Extraction}
In the previous section, we described the numerical features selected, but the process of extracting these features is still ongoing. To extract the features, we used $LLM_E$ models that can be used for the Conditional Generation Task. Notably, in our case, it is a force decoding since the tokens of \emph{generated-text} were potentially generated by a different LLM ($LLM_G$). Instead of letting the model generate the answer token-by-token from the \emph{condition-text} alone, we provide it with the token predicted by $LLM_G$ at each step. This way, $LLM_E$ is forced to follow the path to generate the \emph{generated-text} and, from there, extract the token probabilities from $LLM_E$ if it would generate that sequence itself. Then, using these token probabilities, we compute the four numerical features previously described.

\subsection{Models Specification}
The classifiers used are a Logistic Regression~\cite{wright1995logistic} (LR) and a Simple Neural Network (SNN). Both for a data point of the type \emph{(condition-text, generated-text)} only use the four numerical features extracted. We selected the LR for its simplicity, fast training, and effectiveness in binary classification tasks. However, we implemented a SNN to explore complex non-linear relationships in the data. The SNN architecture consists of an input layer with four neurons representing each features, followed by two hidden layers, each comprising 512 neurons with the ReLU activation function. Followed by an output layer, containing a single neuron activated by a sigmoid function, suitable for binary classification tasks.



\section{Experimental Setup and Results}

In this section, we describe the details of our experimental setup, the dataset we use, and the results we obtain. 

\subsection{Datasets}
Here, we list the three datasets we are using for experimentation and comparison.

\begin{itemize}[noitemsep]
    \item \textbf{HaluEval}: Hallucination Evaluation for Large Language Models (HaluEval) benchmark is a collection of generated and human-annotated hallucinated samples for evaluating the performance of LLMs in recognizing hallucinations. HaluEval includes 5,000 general user queries with ChatGPT responses and 30,000 task-specific examples (10,000 per task) from three tasks: question answering, knowledge-grounded dialogue, and text summarization~\cite{li-etal-2023-halueval}. 
    \item \textbf{HELM}: Hallucination detection Evaluation for multiple LLMs (HELM) benchmark is a list of 3582 sentences from randomly sampling 50,000 articles from WikiText-103~\cite{merity2016pointer} where the selected LLMs were tasked with prompt-based continuation writing. The resulting sentences were annotated as hallucination or not~\cite{su2024unsupervised}. 
    \item \textbf{True-False}: Comprises 6,084 sentences divided into the topics of ``Cities," ``Inventions," “Chemical Elements," “Animals," “Companies," and “Scientific Facts." All sentences in each category are conformed of true statements and false statements~\cite{azariamitchell2023internal}. Unlike HaluEval, this dataset only has \emph{generated-text} and does not include any \emph{condition-text}.
\end{itemize}





\subsection{LLM Evaluators Used}


The LLMs selected as evaluators to study the impact of factors such as architecture, training method, and training data include \emph{GPT-2}, its large version (\texttt{gpt2-large})~\cite{radford2019language}; Bidirectional and Auto-Regressive Transformers (\emph{BART}), its CNN-Large version (\texttt{bart-large-cnn})~\cite{lewisetal2020bart}; Longformer Encoder-Decoder (\emph{LED})~\cite{beltagy2020longformer}, with the version fine-tuned on the arXiv dataset (\texttt{led-large-16384-arxiv}). Also, we used four bigger LLMs like OPT-6.7B (\emph{OPT})~\cite{zhang2022opt}, GPT-J-6.7B (\emph{GPT-J})~\cite{wang2021gpt}, LLaMA-2-Chat-7B (\emph{LLC-7b})~\cite{touvron2023llama} and Gemma-7b (\emph{Gemma})~\cite{team2024gemma}. We used the known transformers library.\footnote{\url{https://huggingface.co/}}
In most cases, we utilized the \texttt{Conditional Generation setup}. For \emph{GPT-2}, we employed the \texttt{GPT2LMHeadModel} setup. Additionally, when forwarding inputs to these models with a pair of \emph{(condition-text, generated-text)}, we encountered the challenge of context limitation, which varied depending on the LLM. To address this issue, we did not truncate the \emph{generated-text} if possible. Instead, if truncation was necessary (with a truncation length of \texttt{truncate\_len}), we removed the excess from the \emph{condition-text}. If additional knowledge was included, we evenly split the truncation between the knowledge and the \emph{condition-text}.


\subsection{Training Process of the Classifiers}

\begin{table*}[!ht]
    \centering	
    \caption{Average results in the test set for each task in the HaluEval benchmark given the selected $LLM_E$. $\text{NC}_\text{Acc}$ stands 
 for accuracy without \emph{condition-text} and $\text{K}_\text{Acc}$ for accuracy including extra knowledge.}
	\begin{tabular}{l|rrrr|rrrrr|rrrrr}
		\toprule
        \multicolumn{1}{c}{\textbf{}} &
		\multicolumn{4}{c}{\textbf{Summarization}} & \multicolumn{5}{c}{\textbf{Question Answering}} & \multicolumn{5}{c}{\textbf{KGD}} \\  
        \cmidrule(r){1-15} 
        \cmidrule(l){5-15}
		Models & Acc & F1 & $\text{PR}_\text{AUC}$ & $\text{NC}_\text{Acc}$ & Acc & F1 & $\text{PR}_\text{AUC}$ & $\text{NC}_\text{Acc}$ & $\text{K}_\text{Acc}$ & Acc & F1 & $\text{PR}_\text{AUC}$ & $\text{NC}_\text{Acc}$ & $\text{K}_\text{Acc}$ \\
		\cmidrule(lr){1-15} \cmidrule(rl){5-15}
  
		GPT-2 & 0.82 & 0.78 & 0.89 & 0.90 & 0.82 & 0.82 & 0.86 & 0.78 & 0.88 & 0.62 & 0.60 & 0.70 & 0.64 & 0.62 \\
  
		BART & 0.77 & 0.78 & 0.83 & \textbf{0.99} & \textbf{0.95} & \textbf{0.95} & 0.97 & \textbf{0.96} & \textbf{0.94}  & \textbf{0.66} & 0.63 & 0.74 & 0.65 & 0.60 \\
  
		LED & 0.97 & 0.76 & 0.81 & 0.97 & 0.88 & 0.88 & 0.91 & 0.86 & 0.87 & 0.62 & 0.61 & 0.70 & 0.62 & 0.60 \\
        
        OPT & \textbf{0.98} & \textbf{0.98} & \textbf{0.99} & \textbf{0.99} & 0.79 & 0.78 & 0.85 & 0.76 & 0.79 & \textbf{0.66} & \textbf{0.67} & \textbf{0.74} & \textbf{0.67} & 0.61 \\

        GPT-J & \textbf{0.98} & \textbf{0.98} & \textbf{0.99} & \textbf{0.99} & 0.77 & 0.78 & 0.84 & 0.73 & 0.83 & \textbf{0.66} & \textbf{0.67} & \textbf{0.74} & \textbf{0.67} & \textbf{0.66} \\
  
		LLC-7b & 0.67 & 0.68 & 0.69 & 0.77 & 0.73 & 0.69 & 0.81 & 0.75 & 0.77 & 0.60 & 0.54 & 0.64 & 0.63 & 0.61  \\

        Gemma & 0.51 & 0.51 & 0.52 & 0.57 & 0.76 & 0.73 & 0.82 & 0.79 & 0.71 & 0.58 & 0.58 & 0.66 & 0.62 & 0.55 \\

		
		\bottomrule
  
  
  
	\end{tabular}
	\label{table:dense_results}
\end{table*}

\begin{table}[t]
\centering
\caption{Results taken from~\cite{li2024dawn} measured in Accuracy (\%) on the Halu-Eval dataset.}
\begin{tabular}{l|rrrr}
\toprule
{Models} & {QA} & {KGD} & {Summ.} & {GUQ} 
\\ \hline
ChatGPT & 62.59 & \textbf{72.40} & \textbf{58.53} & 79.44
\\ 
Claude 2 & \textbf{69.78} & 64.73 & 57.75 & 75.00
\\ 
Claude & 67.60 & 64.83 & 53.76 & 73.88
\\ 
Davinci-003 & 49.65 & 68.37 & 48.07 & 80.40
\\ 
Davinci-002 & 60.05 & 60.81 & 47.77 & \textbf{80.42}
\\ 
GPT-3 & 49.21 & 50.02 & 51.23 & 72.72
\\ 
Llama-2-Ch & 49.60 & 43.99 & 49.55 & 20.46
\\ 
ChatGLM 6B & 47.93 & 44.41 & 48.57 & 30.92
\\ 
Falcon 7B & 39.66 & 29.08 & 42.71 & 18.98
\\ 
Vicuna 7B & 60.34 & 46.35 & 45.62 & 19.48
\\ 
Alpaca 7B & 6.68 & 17.55 & 20.63 & 9.54
\\ \bottomrule
\end{tabular}
\label{tab:previous_results}

\end{table}


\begin{table}[t]
\centering
\caption{Our results for each $LLM_E$ and task using the LR classifier and measure in accuracy on the test set.}
\begin{tabular}{l|rrrrr}
\toprule
{Model} & {Summ.} & {QA} & {KGD} & {GUQ} 
\\ \hline
GPT-2 & 0.66 & 0.77 & \textbf{0.62} & 0.77
\\ 
BART & 0.65 & \textbf{0.94} & 0.49 & 0.54
\\ 
LED & 0.55 & 0.87 & \textbf{0.62 }& 0.52
\\ 
OPT & 0.73 & 0.75 & 0.61 & 0.80
\\ 
GPT-J & \textbf{0.90} & 0.75 & 0.61 & \textbf{0.81}
\\ 
LLC-7b & 0.70 & 0.74 & 0.55 & \textbf{0.81}
\\ 
Gemma & 0.52 & 0.73 & 0.53 & 0.80
\\ \bottomrule
\end{tabular}
\label{tab:lgr_results}

\end{table}


To train LR, we used the sklearn library\footnote{\url{https://pypi.org/project/sklearn/}} with the Limited-memory Broyden-Fletcher-Goldfarb-Shanno Algorithm solver~\cite{saputro2017limited} and default parameters. The SNN was trained during $10^4$ epochs. We used the Adam~\cite{kingma2014adam} optimizer with learning rate of $10^{-3}$. All experiments, including feature extraction, training, and evaluation of the classifiers, were conducted on an NVIDIA L40S GPU core with 48GB of memory. Given a dataset and a single $LLM_E$, training takes anywhere from 30 minutes (HELM) to 4.5 hours (HaluEval), depending on the size of the training data and the length of the \emph{condition-text} and \emph{generated-text}.



\subsubsection{HaluEval}
We train both classifiers for each of the tasks in the HaluEval benchmark. Each data point is split into two data points: \emph{(condition-text, right-answer)} and \emph{(condition-text, hallucinated-answer)}. Therefore, the datasets would be of 20,000 examples for each of the Question Answering (QA), Knowledge-Grounded Dialogue (KGD), and Summarization tasks where in each case half of the dataset is comprised of data points of the type \emph{(condition-text, right-answer)} and the other half are of the type \emph{(condition-text, hallucinated-answer)}. In the case of the General-User Queries, the dataset is already in that format, with each data point classified as a hallucination. Therefore, the dataset size is the same, which is 5,000.


Then, with this adaptation of the HaluEval benchmark dataset when we were approaching a given task, we will sample 10\% of the data points (half with the \emph{right-answer} and the other with a \emph{hallucinated-answer}) randomly. These 10\% data points are used to train both classifiers, and we test the model capabilities on the remaining 90\% of the dataset for a given task.

\subsubsection{HELM}
In the HELM dataset~\cite{su2024unsupervised} the sentences were separated into categories depending on which LLM generated them. Therefore, when we wanted to evaluate the sentences generated by a given LLM like \texttt{LlaMa-Chat-7b} (LLC-7b), we would train on all other sentences produced by other LLMs.

\subsubsection{True-False}
We followed the same process as Azaria et al.~\cite{azariamitchell2023internal}. We pick a category like ``animals" for testing and train in all other categories.


\subsection{Results}

\subsubsection{HaluEval}

We evaluate each classifier trained on the 10\% data of the given task on the other 90\%. We selected the following metrics: Accuracy, $F_1$, and Precision-Recall Area Under Curve ($\text{PR}_\text{AUC}$). Table~\ref{tab:previous_results} shows current state-of-the-art results in HaluEval. All the results were extracted from~\cite{li-etal-2023-halueval, li2024dawn}, which is the paper that introduces the HaluEval benchmark and an empirical study from the same authors on factuality hallucination in LLMs. Next, Table~\ref{tab:lgr_results} shows the accuracy results on the test set for each task using every $LLM_E$ selected and the Logistic Regression as the classifier. As it can be appreciated, the Logistic Regression obtains great results compared to what previous approaches would have gotten on the 90\% of the dataset. Finally, Table~\ref{table:dense_results} shows our average~\footnote{Average results with data sampled randomly in three runs.} results per model of our approach in each metric evaluated on the test set for each of the tasks of Summarization, QA, and KGD respectively. 





The methods of Table~\ref{tab:previous_results} are based on In-Context-Learning approaches and evaluated in 100\% of the dataset. In contrast, our approach utilizes supervised learning, but we believe it's fair to compare it to existing methods. We train our models using only 10\% of the data, reserving the remaining 90\% for testing. We argue that the performance of current approaches on the 90\% test set will not deviate significantly from their performance on the full dataset, especially when there's a significant accuracy difference. Consequently, we've chosen not to present our results alongside the baseline results in Table~\ref{tab:previous_results}. For an exact comparison, we would need to apply their methods to the same test set we've used but we do not have the resources for such heavy computation. The key findings in the results showed in Tables~\ref{tab:lgr_results} and Table ~\ref{table:dense_results} are summarized as follows: 

First, our classifiers trained on only 10\% of the data demonstrate effectiveness on the test set (the other 90\%) for the Summarization and Question Answering tasks using as $LLM_E$ the \emph{GPT-J} and \emph{BART} respectively. The best results in both tasks have an accuracy and $F_1$ over 98\% with the SNN classifier and over 90\% with the LR classifier in Summarization and $F_1$ over 95\% in QA. These results outperform the results previous approaches would have gotten on the same test set. 

Additionally, while not surpassing the state-of-the-art, we achieved competitive results in the dialogue task. Finally, on the GUQ task, a table was not included for the SNN classifier since we obtained similar results for each $LLM_E$. When employing various $LLM_E$ models with the SNN classifier, results suggest overfitting to the negative class, yielding an accuracy of 81\%, $F_1$ of 1\%, and $\text{PR}_\text{AUC}$ of 10\%. This can be attributed to data imbalance, where only about 20\% are not hallucinations. An alternative attempt with a training set of 500 positive and 500 negative examples tested on the remaining 4,000 revealed little success, with a best accuracy at 69\% and $F_1$ at 0.23\%.

Also, in the tasks of QA and KGD, our method, including the knowledge in the \emph{condition-text}, improved the accuracy, while in a few, it did not. Adding knowledge can help LLM evaluators provide meaningful token probabilities for any task with our approach.

An unexpected finding we encountered was that when the $LLM_E$ only provided probabilities for the \emph{generated-text} without the \emph{condition-text}, it yielded remarkably high results in Summarization and QA tasks with specific $LLM_E$ models like \emph{GPT-J}, achieving up to 99\% accuracy or \emph{BART} with 96\%. This anomaly prompted us to verify that we had not inadvertently used testing data for training or made similar errors. However, this was not the case and supported by the fact that it did not happen in KGD and GUQ, using the same code, nor did it occur with all $LLM_E$ models in Summarization and QA. We hypothesize that there may be a probabilistic pattern in the Summarization and QA tasks within the HaluEval dataset generation process that our approach can learn with some $LLM_E$. Note that this observation does not render the benchmark useless; instead, it suggests that a supervised approach may not be the most suitable fit, and instead, unsupervised approaches might be more appropriate for evaluation in this benchmark.


Despite the significant improvement achieved by selecting specific $LLM_E$ models in certain tasks, using other $LLM_E$ models still yielded competitive results, with some instances even surpassing state-of-the-art benchmarks. For example, \emph{OPT} coupled with the SNN classifier achieved an $F_1$ score of 79\% in the QA task, while \emph{GPT-2} attained an 82\% $F_1$ score.

\subsubsection{HELM}
\begin{table}[b]
    \centering	
 \caption{Results of our approach and previous methods in the HELM benchmark measured in $\text{PR}_\text{AUC}$.}
    \begin{tabular}{p{4em}|p{2.5em}|p{2.6em}|p{2em}|p{2em}|p{2em}|p{2em}}
		\toprule
        \cmidrule(r){1-7} 
        \cmidrule(l){5-7}

Baselines & Falcon & GPT-J & LLB -7B & LLC -13B & LLC -7B & OPT \\
\cmidrule(lr){1-7} \cmidrule(rl){5-7}

PE-max & 0.648 & 0.750 & 0.685 & 0.444 & 0.493 & 0.726 \\

SCG-NLI & 0.685 & 0.868 & 0.764 & 0.583 & 0.657 & 0.810 \\

\bottomrule

SAPLMA & 0.513 & 0.699 & 0.578 & 0.305 & 0.407 & 0.621 \\


MIND & \textbf{0.790} & \textbf{0.877} & \textbf{0.788} & 0.604 & \textbf{0.676} & \textbf{0.884} \\

\bottomrule
\textbf{Ours} \\
\bottomrule

GPT-2 & 0.683 & 0.847 & 0.759 & 0.618 & 0.616 & 0.850 \\

BART & 0.710 & 0.828 & 0.695 & 0.569 & 0.568 & 0.825 \\

LED & 0.683 & 0.809 & 0.722 & 0.527 & 0.548 & 0.829 \\

OPT & 0.719 & 0.839 & 0.773 & 0.634 & 0.637 & 0.864 \\

GPT-J & 0.702 & 0.808 & 0.751 & \textbf{0.642} & 0.588 & 0.834 \\

LLC-7b & 0.727 & 0.855 & 0.785 & 0.563 & 0.644 & 0.842 \\

Gemma & 0.738 & 0.850 & 0.786 & 0.601 & 0.651 & 0.843 \\

  
  

        


        
        

        

        
        
        
  
  
        \bottomrule
  
	\end{tabular}
\label{table:dense_results_helm}

\end{table}

  
  
  
        

  
  


\begin{table*}[h]
    \centering	
    \caption{Feature importance based on accuracy for three tasks in the HaluEval benchmark given three $LLM_E$.}
	\begin{tabular}{cccc|ccc|ccc|ccc}
		\toprule
        \multicolumn{4}{c}{Features} &
		\multicolumn{3}{c}{Summarization} & \multicolumn{3}{c}{Question Answering} & \multicolumn{3}{c}{KGD} \\  
        \cmidrule(r){1-13} 
        \cmidrule(l){5-13}
		mtp & avgtp & Mpd & mps & GPT-J & BART & LLC-7b & GPT-J & BART & LLC-7b & GPT-J & BART & LLC-7b \\
		\cmidrule(lr){1-13} \cmidrule(rl){5-13}
  
                \checkmark & \checkmark & \checkmark & \checkmark & \textbf{0.98} & 0.77          & \textbf{0.69} & \textbf{0.76} & \textbf{0.95} & \textbf{0.74} & \textbf{0.66} & 0.65          & 0.60          \\
                \checkmark &            &            &            & 0.50          & \textbf{0.79} & 0.50          & 0.64          & 0.92          & 0.50          & 0.56          & 0.60          & 0.50          \\
                           & \checkmark &            &            & \textbf{0.98} & 0.64          & 0.57          & 0.69          & \textbf{0.95} & 0.51          & 0.62          & \textbf{0.66} & 0.50          \\
                           &            & \checkmark &            & 0.50          & 0.61          & 0.54          & 0.72          & 0.90          & 0.73          & 0.62          & 0.58          & \textbf{0.61} \\
                           &            &            & \checkmark & 0.51          & 0.62          & 0.60          & 0.62          & 0.64          & 0.57          & 0.53          & 0.53          & 0.52          \\
  
		\bottomrule
  
	\end{tabular}
\label{table:halu_eval_ablation}

\end{table*}

Table~\ref{table:dense_results_helm} shows the results of our approach in the HELM benchmark~\cite{su2024unsupervised}. The overall results showed that our approach did not surpass MIND~\cite{su2024unsupervised} except with LLC-13B sentences. However, we only use four features, and still, our approach surpasses the results of other methods like SAPLMA~\cite{azariamitchell2023internal} and, in some cases, SelfCheckGPT with Natural Language Inference (SCG-NLI)~\cite{manakuletal2023selfcheckgpt} and others reported in~\cite{su2024unsupervised}. Also, we showed in the Appendix section how removing the \emph{condition-text} affects our approach, which, different from the Halu-Eval, in this dataset, decreases performance as expected.

Still, unlike MIND, which gets its training data unsupervised without annotation, we are training with the annotated data they provided in their HELM benchmark.

\subsubsection{True-False}
The results obtained in this dataset using any of our selected LLMs were below the baseline provided by~\cite{azariamitchell2023internal} and significantly lower than their approach. This highlights a major weakness in our methodology and points out the importance of utilizing hidden layers as features. We recommend that any future method demonstrate its performance on this challenging dataset. Detailed results are provided in the Appendix section.

\subsubsection{Overall Conclusions from Results}

In general, our results demonstrate that our supervised learning approach, utilizing only four features, exhibits competitive performance compared to current methods. Our approach surpasses the current state-of-the-art methods across various scenarios in tasks and datasets.


In the HaluEval benchmark, where generations originate from ChatGPT powered by GPT-3.5, we found that some of the top-performing LLMs were two GPT-based models (\emph{GPT-2} and \emph{GPT-J}), which are the nearest to the LLM-Generator that we were able to test. Interestingly, even models not based on the LLM-Generator, such as \emph{OPT}, \emph{BART}, \emph{LED}, achieved comparable results and surpassed them in some tasks. Notably, smaller models like \emph{BART} outperformed all LLM evaluators in the QA task, suggesting that varying LLMs, regardless of size, can yield superior results due to training data and architecture differences.



In the HELM benchmark, where each test set comprised sentences generated by accessible LLMs, we explored the impact of using the actual LLM-Generator as the LLM-Evaluator (like the case of \emph{GPT-J} or \emph{LLC-7B}) compared to using a different one. Results revealed that different LLMs as evaluators often yielded similar or better results than the corresponding LLM-Generator, showing the advantages of employing diverse LLMs for evaluation purposes. Furthermore, our experiments demonstrated that the performance disparity between larger LLMs like \emph{GPT-J} and \emph{LLC-7B} versus smaller ones such as \emph{GPT-2} and \emph{BART} is not significant in many scenarios.

Finally, in the True-False dataset, it became evident that our method exhibits weaknesses when applied to this type of data, and the features lack the necessary significance for detecting hallucinations present in that dataset.

\subsection{Feature Importance Analysis - Ablation}


We performed experiments using single numerical feature to determine which features were significant or not to the results. Table~\ref{table:halu_eval_ablation} showed for three tasks in HaluEval and three $LLM_E$ how the results in accuracy were affected by which features were used or not. In most cases, the use of all features provides the best results. Then, when each feature was used alone, we discovered that the most meaningful features in our approach were \emph{mtp} and \emph{avgtp}, especially in Summarization and QA. However, in the KGD task, it was more important for bigger models like \emph{GPT-J} and \emph{LLC-7b}, the feature introduced by this paper: $Mpd$.

Additionally, in the Appendix, we conduct a feature analysis using the coefficients obtained from the LR classifier.

\section{Conclusions and Future Works}

This work introduced a supervised learning approach for identifying hallucinations in conditional text generated by LLMs. Leveraging just four features derived from conditional token probabilities, our approach demonstrated competitive performance compared to existing methodologies across various tasks and datasets. Future work includes exploring hybrid methods that combine In-Context-Learning approaches with probabilistic-based methods, including supervised classifiers.


Through extensive evaluation across three datasets, we uncovered insights into the effectiveness of our approach. Our exploration of using different LLMs as evaluators further emphasized the advantages of employing diverse models for evaluation purposes. By comparing results obtained from using the actual LLM-Generator as the evaluator against those from different LLMs, we showcased the potential for alternative models to yield comparable or even superior evaluation outcomes. In future work, it is possible to investigate advanced ensemble learning techniques to further enhance the performance of hallucination detection systems by effectively combining predictions from multiple LLM evaluators. Furthermore, we identified weaknesses in our approach when applied to datasets like the True-False dataset. Future research could explore whether augmenting our four features with hidden layer features could improve the state-of-the-art performance on that dataset.


This research extends to every domain relying on LLMs. By enhancing the reliability of LLM outputs, the proposed method contributes to the ethical use of these models in sensitive applications, such as medical, legal, educational, and financial domains. This work is a step toward creating a reliable method to detect hallucinations in LLMs based on token and vocabulary probabilities.

\section*{Limitations}
The first limitation is the numerical features and models selected as $LLM_E$. While our current approach has demonstrated effectiveness in specific tasks, it may only capture the richness and complexity of some textual content types. The derived features must be more meaningful for tasks like Knowledge-Grounded Dialogue (KGD), which involve intricate context and real-time exchanges. 

Our method outperformed state-of-the-art tasks like Summarization and Question Answering in HaluEval. However, in KGD and General User Queries, it achieved competitive but not leading results. This could hint at potential over-specialization or the need for task-specific feature engineering. Another reason could be the inherent limitations of the LLMs selected as $LLM_E$. 


Additionally, because of the context length limitation of some of the LLMs, we needed to truncate the \emph{condition-text} or extra knowledge, which might cause us to lose the necessary context to get the correct token probabilities to classify correctly. However, some of the LLMs needed more context length to use everything without losing information.

One of the main limitations is that our approach's results and effectiveness may be tied to the characteristics of the datasets used. The model's performance could be skewed if the dataset has inherent biases or lacks diversity in certain aspects. For instance, it might be in the specific patterns obtained on the HaluEval benchmark that these four numerical features are good indicators for detecting this type of hallucination. However, it does not change the fact that current complex state-of-the-art approaches have yet to show this level of performance under the same circumstances.

Although we achieved competitive results in the HELM benchmark, it is worth noting that, apart from SALPMA, most other approaches relied on unsupervised methods. As our method is supervised, it inherently depends on curated and annotated data, which poses a limitation. Additionally, the performance in the True-False dataset highlights a weakness in our approach, suggesting that it may need to be augmented with additional features to be effective in a broader range of scenarios.


Finally, this method is grounded in binary classification. In real-world scenarios, hallucination might be more nuanced, with varying degrees of severity, which our current approach might not account for. Furthermore, there needs to be more interpretability; even when we can get intuition from the numerical features, we cannot obtain the exact explanation of what specific wrong fact or fictitious information is being added. We intend to explore other ideas on datasets that make this separation to increase the interpretability.

\bibliography{refs}

\appendix

\section{Appendix}
\label{sec:appendix}

\begin{table}[b!]
    \centering	
    \caption{Results of our approach and previous methods in the True-False dataset measured in accuracy. The SALPMA results shown are using the 16th hidden layer with LLC-7b.}
	\begin{tabular}{p{5.5em}|p{2.5em}|p{2.5em}|p{2.5em}|p{2.5em}|p{2.5em}|p{2.5em}}
		\toprule
        \cmidrule(r){1-7} 
        \cmidrule(l){5-7}
		\textbf{Model} & \textbf{Cities} & \textbf{Invent.} & \textbf{Elem.} & \textbf{Anim.} & \textbf{Comp} & \textbf{Facts} \\
		\cmidrule(lr){1-7} \cmidrule(rl){5-7}
  
		BERT-5-shot & 0.5416 & 0.4799 & 0.5676 & 0.5643 & 0.5540 & 0.5148 \\
  
		SAPLMA & \textbf{0.9223} & \textbf{0.8938} & \textbf{0.6939} & \textbf{0.7774} & \textbf{0.8658} & \textbf{0.8254}  \\
        
        \bottomrule
        \textbf{Ours} \\
        \bottomrule
            GPT-2       & 0.4312          & 0.5353          & 0.4924          & 0.4920          & 0.5041          & 0.5049          \\
            BART        & 0.3846          & 0.5365          & 0.5172          & 0.4920          & 0.4550          & 0.4607          \\
            LED         & 0.4985          & 0.4954          & 0.5182          & 0.5357 & 0.5191          & 0.4787          \\
            OPT    & 0.4950          & 0.5479 & 0.5118          & 0.4573          & 0.5050          & 0.5392          \\
            GPT-J    & 0.5023          & 0.5308          & 0.5268 & 0.4871          & 0.5283 & 0.5408          \\
            LLC-7b      & 0.5182 & 0.5216          & 0.5267          & 0.5287          & 0.5208          & 0.5669 \\
            Gemma       & 0.5091          & 0.5205          & 0.4870          & 0.4692          & 0.4983          & 0.4705          \\
            
  
        \bottomrule
  
	\end{tabular}
	\label{table:true-false}
\end{table}

\begin{table}[b!]
    \centering	
    \caption{Results of our approach and previous methods in the HELM benchmark measured in $\text{PR}_\text{AUC}$ without \emph{condition-text}.}    
    \begin{tabular}{p{4em}|p{2.5em}|p{2.5em}|p{2.5em}|p{2.5em}|p{2.5em}|p{2.5em}}
		\toprule
        \cmidrule(r){1-7} 
        \cmidrule(l){5-7}
		\textbf{Baselines} & \textbf{Falcon} & \textbf{GPT-J} & \textbf{LLB-7B} & \textbf{LLC-13B} & \textbf{LLC-7B} & \textbf{OPT} \\
		\cmidrule(lr){1-7} \cmidrule(rl){5-7}
  
		PE-max & 0.6479 & 0.7497 & 0.6851 & 0.4439 & 0.4931 & 0.7263 \\
  
		SCG-NLI & 0.6846 & 0.8680 & 0.7644 & 0.5834 & 0.6565 & 0.8103   \\

        \bottomrule
        
		SAPLMA & 0.5128 & 0.6987 & 0.5777 & 0.3047 & 0.4066 & 0.6212  \\


        MIND & \textbf{0.7895} & \textbf{0.8774} & \textbf{0.7876} & 0.6043 & \textbf{0.6755} & \textbf{0.8835}  \\
        
        \bottomrule
        \textbf{Ours} \\
        \bottomrule
        
            GPT-2        & 0.7110          & 0.8097          & 0.7384          & 0.5194          & 0.6085          & 0.7994          \\
            BART         & 0.6853          & 0.8129          & 0.7139          & 0.5624          & 0.5686          & 0.8258          \\
            LED          & 0.7017          & 0.8424 & 0.6931          & 0.5194          & 0.5494          & 0.8204          \\
            OPT     & 0.7163 & 0.7748          & 0.6695          & 0.5608          & 0.6751 & 0.7773          \\
            GPT-J     & 0.7051          & 0.7873          & 0.6984          & \textbf{0.6121} & 0.5856          & 0.7989          \\
            LLC-7b       & 0.6968          & 0.8403          & 0.7423 & 0.5464          & 0.6370          & 0.8395 \\
            Gemma        & 0.6872          & 0.8256          & 0.7319          & 0.5788          & 0.6428          & 0.8265          \\
  
  
        \bottomrule
  
	\end{tabular}
\label{table:dense_results_helm_not_conditioned}

\end{table}

\begin{table*}[t!]
    \centering	
    \caption{Scaled Logistic Regression coefficients assigned to each input feature based on the selected $LLM_E$.}
	\begin{tabular}{l|cccc|cccc|cccc}
		\toprule
        \multicolumn{1}{c}{\textbf{}} &
		\multicolumn{4}{c}{\textbf{Summ.}} & \multicolumn{4}{c}{\textbf{QA}} & \multicolumn{4}{c}{\textbf{KGD}} \\  
        \cmidrule(r){1-13} 
        \cmidrule(l){5-13}
		\textbf{Models} & \textbf{mtp} & \textbf{avgt} & \textbf{Mpd} & \textbf{mps} & \textbf{mtp} & \textbf{avgt} & \textbf{Mpd} & \textbf{mps} & \textbf{mtp} & \textbf{avgt} & \textbf{Mpd} & \textbf{mps} \\
		\cmidrule(lr){1-13} \cmidrule(rl){5-13}
  
		GPT-2 & 1.0 & \textbf{127.8} & 0.69 & 17.69 & 1.01 & 1.0 & 0.01 & \textbf{957.36} & 1.0 & 1.1 & 0.03 & \textbf{15.88}\\
  
		BART & 0.14 & 1.04 & \textbf{51.28} & 0.017 & 0.005 & 0.0017 & 1.49 & \textbf{0.48}  & 1.5 & 0.003 & \textbf{4.4} & 2.03 \\
  
		LED & 0.05 & 0.81 & 0.61 & \textbf{1.34} & 1.0 & 1.26 & 0.007 & \textbf{31.0} & \textbf{24.86} & 0.004 & 0.1 & 2.9 \\

        OPT & 1.0 & \textbf{1791.86} & 0.93 & 70.01 & 1.0 & 1.0 & 0.006 & \textbf{232.66} & 1.0 & 3.02 & 0.02 & \textbf{3.41} \\

        GPT-J & 1.0 & \textbf{1956.69} & 0.93 & 3.45 & 1.1 & 1.0 & 0.006 & \textbf{223.04} & 1.0 & 3.38 & 0.03 & \textbf{4.6}  \\
  
		LLC-7b & 1.0 & \textbf{3.03} & 1.1 & 0.03 & 1.0 & 11.3 & 0.002 & \textbf{22.97} & 1.0 & 0.85 & 0.03 & \textbf{3.94}  \\

        Gemma & 1.0 & 2.36 & 1.0 & \textbf{5.75} & 1.0 & 4.7 & 0.0004 & \textbf{21.5} & 1.0 & 0.86 & 0.1 & \textbf{3.29} \\

		
		\bottomrule
  
	\end{tabular}
	\label{table:lr_coefficients}
\end{table*}

\subsection{True-False Dataset results}
In this appendix section, Table~\ref{table:true-false} provides the exact numerical results of our approach in the True-False dataset, which, as discussed in the paper, yielded significantly weak results. Once again, we want to highlight that while this paper demonstrates competitive and achievable performance on many tasks of HaluEval and HELM using only four features, the performance in this dataset was notably poor. Therefore, we strongly recommend that future hallucination evaluations with supervised approaches be conducted on a dataset like this or on a dataset where a rapid method like ours has been tested to ensure no clear probabilistic pattern.

\subsection{HELM results without condition-text}

In this section, Table~\ref{table:dense_results_helm_not_conditioned} presents the results of our approach in the HELM benchmark using the token probabilities obtained solely from the \emph{generated-text}. We observe how the performance was impacted when using most of the $LLM_E$ models in comparison with the results in Table~\ref{table:dense_results_helm}. This highlights the significance of the \emph{condition-text} and once again suggests that the anomaly observed in the HaluEval dataset in the Summarization and QA tasks may be attributed to a probabilistic pattern in the \emph{generated-text}.

\subsection{Feature Importance Analysis - Logistic Regression Coefficients}

Additionally, we extracted the Logistic Regression coefficients for each feature in each task. Due to the utilization of the logit function, the coefficients in logistic regression signify the logarithm of the odds that an observation belongs to the target class (``1") based on the values of its input variables. Therefore, to interpret these coefficients appropriately, they must be transformed into regular odds. We achieved this by exponentiating the logarithmic odds coefficients, a task easily accomplished using the \texttt{np.exp()} function.

The interpretation in this case can be read as: for each incremental unit rise in the given input variable (for example, the feature $avgtp$), the likelihood of the observation belonging to the positive class increases by a factor of the value of the coefficient, while maintaining all other variables constant, compared to the odds of it being in the negative class.

Table~\ref{table:lr_coefficients} indicates that for the LR classifier, the most significant features during Summarization were predominantly \textbf{avgtp}, but also occasionally included \textbf{Mpd} and \textbf{mps}. However, in the QA task, the most relevant feature consistently remained \textbf{mps}. Similarly, in the KGD task, \textbf{mps} was predominantly critical, but occasionally \textbf{mtp} and \textbf{Mpd} also played a role. However, it's important to note that LR does not achieve results as strong as the SNN. Therefore, the feature importance of SNN, as shown in Table~\ref{table:halu_eval_ablation}, may carry more weight.

\end{document}